\pdfoutput=1

\documentclass[11pt]{article}

\usepackage{ACL2023}

\usepackage{times}
\usepackage{latexsym}

\usepackage[T1]{fontenc}

\usepackage[utf8]{inputenc}

\usepackage{microtype}

\usepackage{inconsolata}

\usepackage{multirow}

\usepackage{graphicx}

\usepackage[inline]{enumitem}
\newlist{inlinelist}{itemize*}{1}
\setlist*[inlinelist,1]{label=\textbullet, itemjoin={{ \ }}}

\usepackage{tasks}

\usepackage[compact]{titlesec}

\usepackage{paralist}

\newcommand{\citeapos}[1]{\citeauthor{#1}'s (\citeyear{#1})}

\title{Toward More Accurate and Generalizable Evaluation Metrics for Task-Oriented Dialogs}

\author{
    Abishek Komma, Nagesh Panyam Chandrasekarasastry, Timothy Leffel\\
    {\bf Anuj Goyal}, {\bf Angeliki Metallinou}, {\bf Spyros Matsoukas}, {\bf Aram Galstyan}\\\\
    Amazon Alexa AI \\\\
    \texttt{\{kommaak,nagecha,leffelt,anujgoya,ametalli,matsouka,argalsty\}@amazon.com}
}

\begin{document}
\maketitle
\begin{abstract}
Measurement of interaction quality is a critical task for the improvement of spoken dialog systems. Existing approaches to dialog quality estimation either focus on evaluating the quality of individual turns, or collect dialog-level quality measurements from end users immediately following an interaction. In contrast to these approaches, we introduce a new dialog-level annotation workflow called Dialog Quality Annotation (DQA). DQA expert annotators evaluate the quality of dialogs as a whole, and also label dialogs for attributes such as goal completion and user sentiment. In this contribution, we show that: ({\it i}) while dialog quality cannot be completely decomposed into dialog-level attributes, there is a strong relationship between some objective dialog attributes and judgments of dialog quality; ({\it ii}) for the task of dialog-level quality estimation, a supervised model trained on dialog-level annotations outperforms methods based purely on aggregating turn-level features; and ({\it iii}) the proposed evaluation model shows better domain generalization ability compared to the baselines. On the basis of these results, we argue that having high-quality human-annotated data is an important component of evaluating interaction quality for large industrial-scale voice assistant platforms.
\end{abstract}

\section{Introduction}
\label{sec:intro}

Automated measurement of interaction quality is a critical task for the development and improvement of large-scale voice-based AI assistants. There has been a substantial amount of recent work on automated dialog evaluation both for open domain (\citealt{ji-etal-2022-achieving}; \citealt{ghazarian2021user}; \citealt{ghazarian-etal-2022-wrong}) and task-oriented (\citealt{bordes2017learning}; \citealt{lubis-etal-2022-reinforcement}) dialog systems (for recent surveys, see \citealt{deriu2021}; \citealt{yeh2021comprehensive}). For task-oriented dialog (TOD) systems such as conversational AI assistants, existing research has largely focused on evaluating the quality of individual turns (\citealt{ultes2014application}; \citealt{schmitt2015interaction}; \citealt{gupta2021robertaiq}; a.o.). Estimating interaction quality at the multi-turn dialog level is a relatively less studied topic (though see \citealt{walker1997paradise}; \citealt{walker2000towards}; \citealt{Bodigutla2020}; \citealt{deng2022user}). 

\citet{bodigutla2019multi}; \citet{Bodigutla2020} showed that user-supplied (binarized) dialog-level satisfaction ratings can be predicted with 70-80\% accuracy on a multi-domain dataset, if turn-level quality predictions are included as input features in a supervised model. However, in real-world scenarios, user-satisfaction ratings can be logistically difficult---and frustrating for users---to gather at a sufficiently large scale. Such ratings are also influenced by factors irrelevant to interaction quality itself, e.g.~novice and expert users might rate the same dialog very differently; see \citealp{bodigutla2019multi} for discussion. 

\begin{figure}[htp]
    \centering
    \scriptsize
    \resizebox{7.5cm}{!}{%
    \begin{tabular}{|p{0.35cm}|p{2cm}|p{4cm}|}
    \hline
        \textbf{Turn} & \textbf{User utterance} & \textbf{System response} \\ \hline
        1 & what is the score of the giants game & the new york giants are trailing the chicago bears 24 to 13 at halftime \\ \hline
        2 & no i meant the san francisco giants baseball team they are playing the cubs & the san francisco giants lost 2 to 0 to the new york yankees last night and their next game is against the chicago cubs at 7 pm eastern tonight. do you want me to remember that the san francisco giants are your favorite baseball team \\ \hline
        3 & no not really thanks though & no problem. did you want a news update for your other favorite sports teams \\ \hline
        4 & sure & here is your sports update. the new york... \\ \hline
    \end{tabular}}
\caption{Example interaction between human end user and conversational AI assistant}
\label{fig:dialog-example}
\end{figure}

Perhaps more importantly, the factors relevant to whether a single dialog turn is ``successful'' (or ``defective'') can be quite different from the factors relevant to whether a multi-turn dialog is successful: even human-to-human linguistic communication frequently involves temporary misunderstandings, clarification, rephrasing, etc.---attributes that are indicative of poor interaction quality {\it only when viewed in isolation}. For example, consider the (synthetic) dialog in Figure \ref{fig:dialog-example}: Even though the system response in Turn 1 does not satisfy the user's request, the user quickly clarifies, and immediately receives relevant information. Furthermore, Turn 2's response shows that the user's initial request was based on an incorrect assumption (that a SF Giants game is underway). Despite this, the system provides enough pertinent information to resolve the original request. Viewed as a whole, this is a high-quality dialog.

In this contribution, we present a scalable approach to dialog-level quality estimation based on a new annotation scheme we call Dialog Quality Annotation (DQA). DQA adapts and extends \citeapos{bodigutla2019domain} turn-level Response Quality (RQ) annotation task to the dialog level. Whereas \citet{bodigutla2019domain} obtain dialog-level quality labels via directly soliciting user-satisfaction ratings, DQA uses expert annotators to collect ground-truth labels.

In line with the results of \citet{bodigutla2019multi}, we found that aggregations of turn-level signals are indeed predictive of dialog-level ratings. However, we also found that a supervised approach utilizing both dialog-level signals and aggregated turn-level signals achieves superior performance (F1=.81) compared to aggregation of turn-level features alone (F1=.73; similar to the findings of \citet{bodigutla2019domain} for predicting single-turn ratings). These results have implications for the design of multi-turn interaction quality measurement systems, chief among which is that such systems will achieve superior performance if they include both features computed over entire dialogs and features derived from individual turns of a dialog.

Our contributions are summarized as follows: 

    1. We develop a high-velocity dialog quality annotation (DQA) scheme and use it to generate dialog-level annotations for 3674 dialogs across 11 different domains.
    
    2. We use the annotated data to train a supervised model for predicting binarized dialog-level quality ratings.
    
    3. We conduct experiments and find that our proposed model outperforms baselines in F1 score, and generalizes better to an unseen domain, thus showcasing the value of high-quality dialog-level annotations.

\section{Related Work}
\label{sec:related}

Existing research on quality metrics for multi-turn human-computer interactions has focused on either task-oriented dialog systems, or open-domain (``chitchat'') systems. The present study concerns largely task-oriented use cases, but given the conversational nature of our platform, chitchat also can (and does) occur in dialogs we evaluate.

\subsection{TOD Systems}
Task-oriented dialog (TOD) systems help humans to achieve concrete tasks via voice or text interaction. For example TOD systems help users book reservations, communicate with customer service systems, or navigate menus. Evaluating the quality of such interactions requires a dataset of TODs annotated with quality scores. A number of TOD datasets have been released publicly (see \S 4.1 of \citealt{sun2021}), but most are designed to evaluate the performance of dialog understanding tasks like Dialog State Tracking, as opposed to the quality of dialogs from the perspective of successful communication. Many such public datasets were created via Wizard-of-Oz experiments, i.e.~human-human interactions where one human plays the role of system and the other of user \citep{eric2019multiwoz}. Other datasets were collected by first simulating dialog outlines in the form of API sequences and then asking annotators to expand the outlines into natural language dialogs \citep{rastogi2020schema}. A recent study annotated TOD datasets with user satisfaction scores by showing dialogs to annotators and asking them to rate for quality \citep{sun2021}.

Various annotation schemas have been proposed to label the quality of TODs at the turn-level. In Interaction Quality (IQ), raters were asked to rate each turn on a 1-5 scale, taking into consideration the dialog quality so far \citep{schmitt2012}. To reduce the cognitive load on annotators, \citet{bodigutla2019domain} proposed the Response Quality (RQ) annotation schema. RQ removed the constraint to keep track of the dialog quality so far, but asked annotators to consider if the next user utterance might contain feedback, such as frustration, rephrasing, etc. The RQ scale is: 1=Terrible (fails to understand user's goal), 2=Bad (understands goal but fails to satisfy it in any way), 3=OK (partially satisfies goal), 4=Good (mostly satisfies goal), and 5=Excellent (completely satisfies user's goal). Another recent study \citep{sun2021} collected annotations at the dialog level, using a simple (underspecified) 5-point user satisfaction scale.

Various approaches have been explored to train models to estimate task-oriented dialog quality. Earlier approaches used text-based features from dialogs and trained models like SVMs to predict quality scores. More recent approaches use RNNs (sometimes hierarchical) or BERT to encode dialogs and train models to predict turn- and/or dialog-level quality scores. These approaches model the task either as classification (for discrete quality scores) or regression (for quantitative quality scores).  
Recent research has explored applications of large language models (LLMs) for dialog-based NLU tasks such as intent recognition and dialog state tracking. Such models have been trained using publicly available TOD datasets, e.g.~\citet{wu2020tod}; \citet{peng2020soloist}; \citet{yang2021ubar}. TOD-based LLMs have not been explored as extensively for the purpose of TOD quality estimation, though this is an active area of research for us.

See \citealt{deriu2021} for a survey of approaches to evaluation in TOD systems.

\subsection{Open-Domain Dialog Systems}
Developing quality metrics for open-domain dialog systems presents different challenges than for TOD systems. In an open-domain dialog, a system can have many relevant responses for a single utterance, and a single dialog could cover multiple unrelated topics. Automated evaluation approaches have explored different aspects of dialog quality such as coherence, informativeness, user engagement~\citep{vakulenko2018measuring, zheng2021dynaeval, mehri2020fed,ghazarian2020engagement}. Similar to TOD, open-domain dialog evaluation requires high-quality training data. Existing work has used datasets by collecting human judgments~\citep{higashinaka2014evaluate, cervone2020coherent}. Another general approach is to use conversations between human users as coherent/positive examples, and then generate negative examples/incoherent dialogs by applying certain perturbations to the coherent dialogues, such as shuffling order or injecting irrelevant utterances into the dialog~\citep{vakulenko2018measuring, mesgar2020dialogue, huang2020grade, zheng2021dynaeval}. Recent work has considered higher-level semantic perturbations that change the dialog flow more subtly~\citep{ghazarian-etal-2022-deam}.

\section{Dialog Quality Annotation}
\label{sec:worksflow}

\subsection{DQA Workflow}
Here we describe the workflow for generating annotations needed to train a supervised dialog quality estimation model. This workflow adapts and extends the related turn-level Response Quality (RQ) workflow of \citet{bodigutla2019multi}. We refer to this workflow as ``Dialog Quality Annotation'' (DQA). DQA is platform- and domain-agnostic, and was designed to support high-velocity annotation.

In each DQA task, a multi-turn dialog is presented in its entirety to an expert data annotator (DA). First, the DA is asked to rate the quality of each turn in the dialog. After each turn has been annotated, the DA then answers questions about the dialog as a whole (overall dialog rating, number of goals, goal completion, goal progression, goal friction, system response coherence, and user's inferred sentiment). DAs assigned quality scores to dialogs using a five-point rating scale. About 20\% of dialogs are annotated by two DAs, for quality control monitoring. After the workflow was fully productionized and DAs were calibrated on the annotation task, we have observed weekly inter-rater agreement rates ranging from 79\% to 86\% (with a difference of one scale point allowed). See Appendix \ref{sec:appendix} for further details about the workflow.

Using the DQA workflow, we gathered a dataset of 3569 annotated dialogs (9347 turns from 3233 unique users), of which 714 were held out as a test set to evaluate the performance of baseline methods and trained models. The remaining 2855 annotated dialogs were used to train candidate dialog-level defect detection models. This data was gathered by randomly sampling (de-identified) interactions across 10 different experiences supported by our platform. Our train-test split was stratified by experience, so that each use case appears at a similar rate across train and test sets.

Finally, we gathered 105 additional annotated dialogs (502 total turns) from a use case that does not appear in the training or test data (Shopping product Q\&A). These out-of-distribution (OOD) dialogs enable us to more realistically assess how well the resulting model generalizes to patterns unseen during training.

The majority of the data we gathered were from experiences in which the system only has access to information about the target use case. Such traffic is partitioned into discrete user sessions by default, so we considered a ``dialog'' to just be a single user session. For the OOD traffic, which does not come with pre-defined session boundaries, we used a time-based heuristic where a dialog is considered to be a sequence of utterances from a single user, with no more than 180 seconds of inactivity between turns. In future work, we are exploring model-based methods for dialog segmentation.

\subsection{Dialog quality versus dialog attributes}

As discussed above, for every dialog, the DAs provide the overall dialog-level rating, salient attributes of the dialog, and the individual turn-level ratings. With these annotations we aim to understand the relationship between salient attributes of a dialog (e.g.~goal progression, goal completion, response coherence) and the overall dialog-level ratings. The motivation here is that a robust relationship between objective dialog attributes and dialog ratings would help us to derive human-quality labels from automated methods in the future. While some research exists on the relationship between turn-level and dialog-level quality ratings (\citealt{Bodigutla2020}), few studies explore the relationship between dialog-level attributes and dialog-level quality ratings (\citealt{siro}).

\begin{figure}[htp]
	\includegraphics[width=\linewidth]{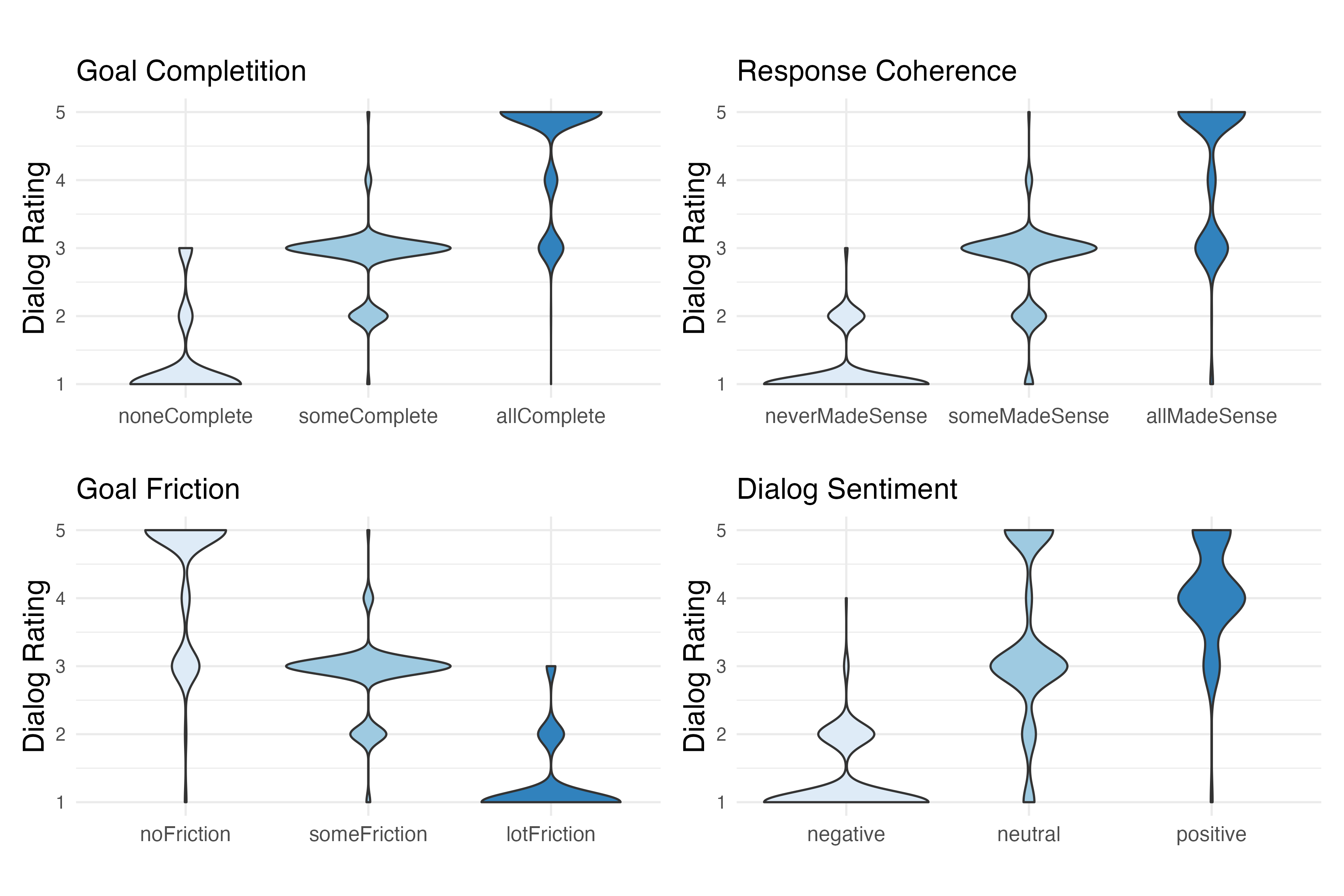}
	\caption{Distribution of dialog ratings with salient attributes of dialog.}
	\label{fig:salient}
\end{figure}

In Figure \ref{fig:salient} we plot the distribution of dialog-level rating against four salient attributes of the dialog. As expected we can clearly see that dialogs received higher ratings when users successfully completed their goals, system responses were coherent, and users encountered less friction while progressing towards their goals. Further, Table \ref{tab:dialog_salient_attributes} computes the Spearman's $\rho$ correlation between the ratings and attributes. Goal completion was found to have the highest correlation score of .859, while user sentiment had the lowest, at .449. Moreover, user friction encountered had a negative correlation to dialog rating. These observations are intuitive given the dialogs were sampled from mostly task-oriented experiences.

\begin{table}[htp]
	\caption{Correlation of dialog rating with salient attributes of the dialog. All correlations in this table are statistically significant at $p < 0.01$.}
	\begin{tabular}{l|r}
		\hline
		~Attribute~  & ~Spearman's $\rho$~ \\ 
		\hline
		Goal Completion    & .859     \\
		Response Coherence & .766     \\
		Goal Friction      & (.807)   \\
		User Sentiment     & .449     \\
		\hline
	\end{tabular}
	\label{tab:dialog_salient_attributes}
\end{table}

\section{Dialog Quality Estimation Model}
\label{sec:DQM-model}

We now describe our dialog quality estimation model (DQM), which leverages the dialog-level annotations described in the previous section.

\begin{figure}[htp]
	\includegraphics[width=\linewidth]{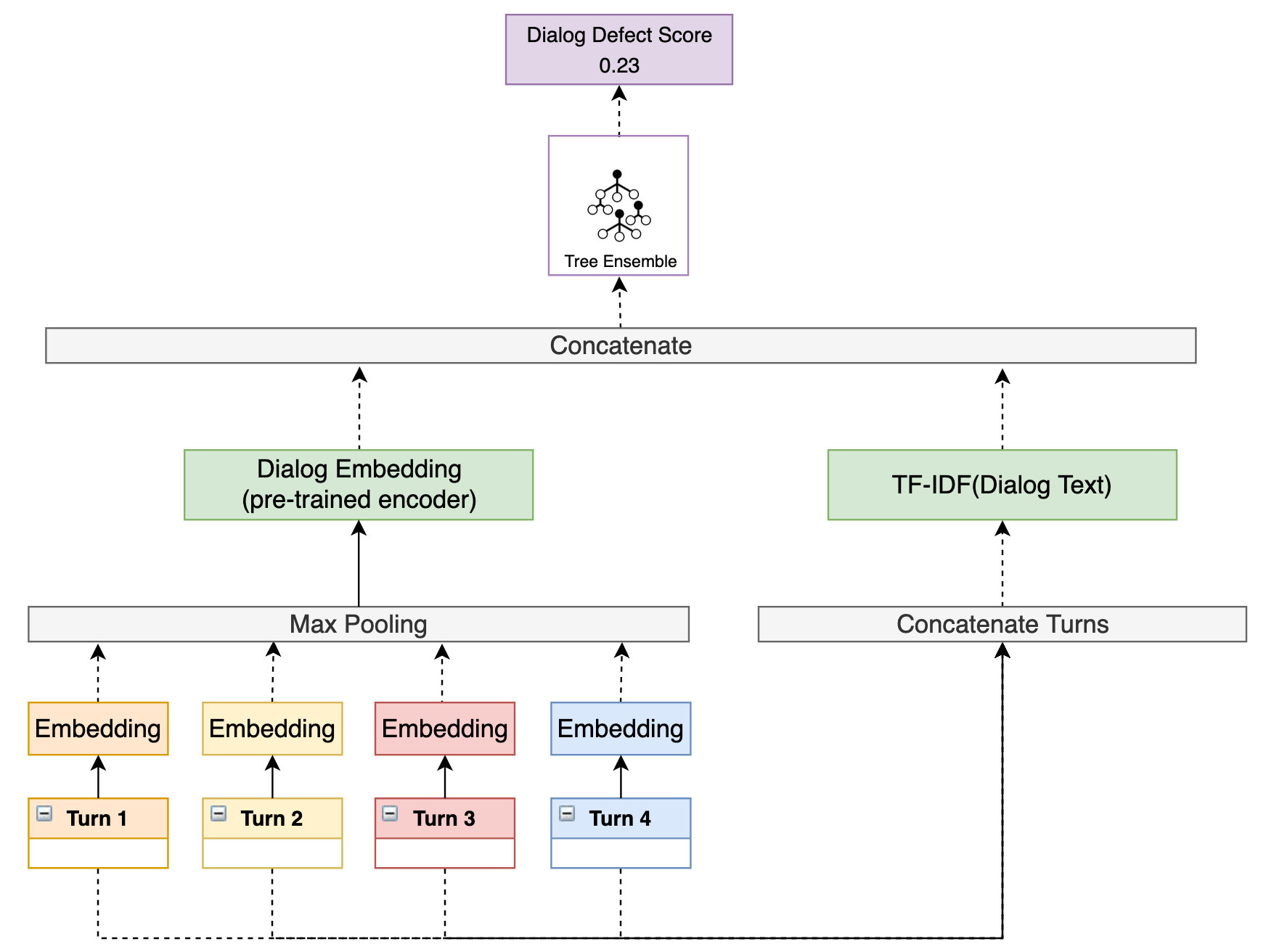}
	\caption{DQM Model Architecture}
	\label{fig:model-arch}
\end{figure}

Figure~\ref{fig:model-arch} illustrates the architecture of the model. We first leverage a pre-trained turn-level defect detection model (which is trained on millions of interactions) as a feature extractor using a RoBERTa-IQ-based framework \citep{gupta2021robertaiq}. We encode each turn of a dialog as a dense vector. We use a max-pooling operation on the turn-level vectors to obtain a dialog-level representation. Finally, we concatenate this with a bag-of-words representation (TF-IDF over unigrams) of the dialog text. This final dialog-level vector is then fed into a Random Forest Classifier to learn a mapping from dialog-level representation to the binarized defect label in $\{0, 1\}$. We arrived at this setup after experimenting with various text and numeric features, and simpler classification algorithms. We also found more sophisticated models to be less effective with our current dataset, although we plan to revisit more complex architectures as the data grows.

Experimental results comparing the performance of this model against several strong baselines are presented in \S\ref{sec:experiments}-\ref{sec:results}. 

The pre-trained text encoder used in our model is based upon an internal model that produces turn-level defect (TLD) scores, which are real-valued scores in $[0,1]$ that can be interpreted as the probability that a given turn is defective from the perspective of the user (see 
\citealp{gupta2021robertaiq} 
for details on the model). TLD scores are derived from a 
RoBERTa-IQ 
classifier trained to detect defective turns within a dialog. Although the TLD model does take context into account when scoring interactions, it is explicitly designed to score dialog {\it turns}, as opposed to entire dialogs.

Our primary question was therefore whether a model trained on the task of dialog-level defect detection outperforms methods that only involve aggregation of turn-level signals. The relevant tradeoff here is that aggregations of TLD scores are cheap and easy to compute, but may suffer from poor accuracy since they were not designed to make predictions about dialogs as a whole.

We hypothesized that a dialog-level statistical model would outperform the TLD-based baselines, in large part because of observed interaction patterns in which the quality of a dialog is not a straightforward combination of the quality of its constituent turns.

\section{Experiment Setup}
\label{sec:experiments}

For the purposes of these experiments, we binarized the five-point dialog-level quality labels by assigning dialogs rated 1, 2, or 3 to the defect class, and dialogs rated 4 or 5 to the non-defect class. This follows the approach taken by \citet{gupta2021robertaiq} for turn-level response quality prediction, enabling us to frame defect detection as a binary classification task.

We assessed the quality of each estimator by measuring its precision, recall, and F1 score relative to human labels on the held-out test set.

We computed four baseline dialog-quality scores, all of which were derived by aggregating TLD scores across each turn in a dialog. We expected to see a very strong relationship between average TLD and dialog quality score, especially since the TLD model uses information from surrounding turns as features. 

These are the baseline methods we computed over the test data used for model evaluation. Each score reduces a sequence of turn-level scores from a dialog into a single value, which represents the dialog-level score.

1. \textit{Mean TLD}: Simple arithmetic mean of the predicted turn-level TLD model scores. 

2. \textit{Last-turn TLD score}: Interpret the final turn's TLD score as the dialog-level score. The idea is that recency bias will lead the final turn to have more impact than others in perceived dialog quality.

3. \textit{Union of mean and last-turn TLD}: A dialog is considered defective if either the mean or last-turn TLD score exceeds some threshold (here: 0.5).

4. \textit{Rising linear weights}: Calculate mean TLD score with each turn linearly weighted by its index, so that later turns have higher weights.

Baseline methods required no training process at all, as they consist of arithmetic aggregations of TLD scores, which were already available prior to experimentation. To prepare each dialog for baseline evaluation, we simply computed each aggregation for each dialog. Baseline aggregations were then converted to binary predictions via a threshold: dialogs with scores $\geq.5$ are considered defective; scores $<.5$ are considered non-defective (we found that some use cases achieve higher accuracy with higher thresholds, while others benefit from lower thresholds; here we use the fixed value of $.5$, as we intend for these methods to be applicable to any supported use case). We scored each dialog in the 714-dialog test set and the 105-dialog OOD test set for each baseline method, and computed performance metrics of interest relative to the human annotations.

To optimize hyperparameters and perform feature selection for our candidate dialog-level defect detection model, we used five-fold cross validation over the training set, selecting the fit that maximized (mean) F1 over the set of hyperparameters and feature subsets considered. The resulting configuration was then trained against the entire 2855-dialog training set. We then used the resulting model to predict defect class (and class probability) over both test sets, computing and recording the same performance metrics of interest.

\begin{table*}[htp]
	\caption{Performance of TLD-based baselines and supervised model}
	\begin{tabular}{l|ccc|ccc}
		\hline
		\multicolumn{1}{c|}{\multirow{2}{*}{}} & \multicolumn{3}{c|}{Multi-domain test set ($n=714$)}  & \multicolumn{3}{c}{OOD test set ($n=105$)} \\
		\multicolumn{1}{c|}{}          & Precision & Recall    & F1-Score  & Precision & Recall    & F1-score  \\ \hline
		Mean TLD                       & {\bf .84} & .54       & .66       & .39       & .77       & .52       \\
		Last-turn TLD                  & .83       & .68       & .75       & .47       & .23       & .31       \\
		Union of mean \& last-turn TLD & .82       & .73       & .77       & .38       & .77       & .51       \\
		Rising linear weights          & .83       & .63       & .72       & .41       & .67       & .51       \\ 
		DQM                            & .78       & {\bf .83} & {\bf .81} & {\bf .48} & {\bf .80} & {\bf .60} \\
		\hline 
	\end{tabular}
	\label{tab:baseline_dqm_metrics}
\end{table*}

\section{Results}
\label{sec:results}
We present the experimental results of the baseline methods and our supervised model for dialog level defect detection (DQM) in Table~\ref{tab:baseline_dqm_metrics}. We describe our observations and inferences from this comparative study in the following section.

\subsection{Performance of baselines and DQM}
\label{subsec:exp-results}

We observed the following regarding the performance of TLD-based baselines and DQM:

1. {\it Among the TLD-based baselines, the Union of Mean \& Last Turn TLD performs best in all scenarios.} However, in absolute terms, the best baseline is not the best performing method for evaluating dialog quality, and only achieves F1 scores of .77 and .51 compared to human annotation on the multi-domain and OOD data, respectively.

2. {\it DQM outperforms the best TLD-based baseline in F1 by 4 and 9 percentage points on the multi-domain and OOD test sets, respectively}. Note that the OOD (Shopping) use case was unseen during training, yet the model achieves an out-of-the-box F1 score of .60 in detecting defective OOD dialogs, compared to only .51 for the best baseline.

3. {\it DQM has a large advantage in recall over baselines, albeit at the cost of reduced precision}.

\subsection{Error analysis}

We further analyzed the performance of DQM and baseline methods over the test set, splitting the data by various attributes of interest. We made the following inferences on the basis of these analyses:

1. {\it Performance of TLD-based baselines and DQM as a function of dialog length indicates that the gap widens as dialog length increases.} Baselines perform better for shorter dialogs ($\leq 3$ turns) and start to drop in performance as dialog length increases, while DQM's performance improves as dialog length increases. This observation likely explains part of the gap between DQM and baselines on OOD data, since these dialogs tend to be much longer than in our multi-domain dataset (mean of $4.78$ turns per dialog versus $2.62$). Table \ref{tab:perf_against_dialog_length} shows baseline versus DQM performance over the multi-domain test set, split by dialog length.

2. {\it DQM has an advantage in detecting defective dialogs that contain a small number of fatal turns}, early on or in the middle of the dialog, which create an overall defective experience. In contrast, TLD-based baselines like mean TLD weight each turn equally and often miss such dialogs. See Appendix~\ref{sec:fatal-turns} for further discussion of this pattern.

3. {\it Both TLD-based baselines and DQM struggle to differentiate between user query rephrasing, which is typically a defect, and user query refinement, which is typically not a defect} (see Appendix \ref{sec:user-rephrase} for examples). User rephrasing happens when a user request is not successful and the user repeats their request with a slightly different surface form. 
User refinement occurs when a user iteratively refines a successful search by adding or modifying constraints. 
We observe that TLD-based baselines have a bias towards incorrectly predicting refinements as defects, possibly because it misclassifies them as rephrases. DQM also struggles with this since it uses TLD as input signal. We hypothesize that these biases may be easier to correct by retraining DQM with targeted multi-turn data than by retraining the TLD model, which is primarily trained on single- or few-turn interaction patterns.

\begin{table}[htp]
	\caption{Performance (ROC-AUC) of TLD-based baselines and supervised model against dialog length on Multi-domain test set ($n=714$). TLD-U is union of mean and last-turn TLD (the best baseline).}
	\begin{tabular}{l|r|rr}
		\hline
		~Dialog Length     & ~$n$~ & ~TLD-U~ & ~DQM~ \\ 
		\hline
		Short ($\leq 3$ turns)  & 535   & .76     & .79   \\
		Medium ($4$-$6$ turns) & 149   & .73     & .80   \\
		Long  ($\geq 7$ turns)  & 30    & .69     & .84   \\
		\hline
	\end{tabular}
	\label{tab:perf_against_dialog_length}
\end{table}

\section{Conclusion}
\label{sec:conclusion}

In this study, we presented a new dialog-level annotation workflow DQA, which enables high-velocity labelling of multi-turn human-computer interactions. Our approach is similar to \citet{Bodigutla2020}, but differs in that we gather labels from expert annotators instead of end users themselves.

We showed that a supervised model trained on DQA annotations outperforms several strong baselines based on aggregating turn-level defect scores. Furthermore, we observed that the model generalizes better to a previously unseen domain. We also found several qualitative patterns of interest, most notably that DQM's advantage over baselines expands as dialog length increases. These findings jointly lend support for an annotation-based approach to estimating multi-turn interaction quality for large-scale dialog systems.

\newpage
\section*{Limitations}
Our proposed approach is designed explicitly for evaluation of task-oriented dialog systems, and is hence unlikely to generalize well to chitchat systems. Most traffic to our platform (and our annotation workflows, including DQA) comes in the form of task-oriented interactions. User turns in the traffic we analyze tend to be quite short (usually less than 20 tokens) and direct, so our model is unlikely to perform as well on dialogs driven by long-form user utterances. 

\section*{Ethical Considerations}
We do not envision any ethical concerns with the research presented here. No customer data is released or presented in this paper, and even our internal data sources are fully de-identified and contain no customer Personal Identifiable Information (PII).

\section*{Acknowledgments}
We wish to thank: Di Wang and Wenbo Yan of Alexa Shopping for providing the OOD test data; the Amazon Data Services team for their work producing annotations; and the metrics team in Alexa for developing the turn-level model that forms the backbone of the dialog-level model presented here. And thanks to the anonymous reviewers, whose feedback helped to clarify and improve this paper.

\bibliography{anthology,custom}

\begin{thebibliography}{34}
\expandafter\ifx\csname natexlab\endcsname\relax\def\natexlab#1{#1}\fi

\bibitem[{Bodigutla et~al.(2019{\natexlab{a}})Bodigutla, Polymenakos, and
  Matsoukas}]{bodigutla2019multi}
Praveen~Kumar Bodigutla, Lazaros Polymenakos, and Spyros Matsoukas.
  2019{\natexlab{a}}.
\newblock Multi-domain conversation quality evaluation via user satisfaction
  estimation.
\newblock \emph{3rd Workshop on Conversation AI: Today’s Practice and
  Tomorrow’s Potential, 33rd Conference on Neural Information Processing
  Systems}.

\bibitem[{Bodigutla et~al.(2020)Bodigutla, Tiwari, Valls-Vargas, Polymenakos,
  and Matsoukas}]{Bodigutla2020}
Praveen~Kumar Bodigutla, Aditya Tiwari, Josep Valls-Vargas, Lazaros
  Polymenakos, and Spyros Matsoukas. 2020.
\newblock \href
  {https://www.amazon.science/publications/joint-turn-and-dialogue-level-user-satisfaction-estimation-on-multi-domain-conversations}
  {Joint turn and dialogue level user satisfaction estimation on multi-domain
  conversations}.
\newblock In \emph{EMNLP 2020}.

\bibitem[{Bodigutla et~al.(2019{\natexlab{b}})Bodigutla, Wang, Ridgeway, Levy,
  Joshi, Geramifard, and Matsoukas}]{bodigutla2019domain}
Praveen~Kumar Bodigutla, Longshaokan Wang, Kate Ridgeway, Joshua Levy, Swanand
  Joshi, Alborz Geramifard, and Spyros Matsoukas. 2019{\natexlab{b}}.
\newblock Domain-independent turn-level dialogue quality evaluation via user
  satisfaction estimation.
\newblock \emph{arXiv preprint arXiv:1908.07064}.

\bibitem[{Bordes et~al.(2017)Bordes, Boureau, and Weston}]{bordes2017learning}
Antoine Bordes, Y-Lan Boureau, and Jason Weston. 2017.
\newblock \href {https://openreview.net/forum?id=S1Bb3D5gg} {Learning
  end-to-end goal-oriented dialog}.
\newblock In \emph{International Conference on Learning Representations}.

\bibitem[{Cervone and Riccardi(2020)}]{cervone2020coherent}
Alessandra Cervone and Giuseppe Riccardi. 2020.
\newblock Is this dialogue coherent? {L}earning from dialogue acts and
  entities.
\newblock In \emph{Proceedings of the 21th Annual Meeting of the Special
  Interest Group on Discourse and Dialogue, SIGdial 2020}, pages 162--174.
  Association for Computational Linguistics.

\bibitem[{Deng et~al.(2022)Deng, Zhang, Lam, Cheng, and Meng}]{deng2022user}
Yang Deng, Wenxuan Zhang, Wai Lam, Hong Cheng, and Helen Meng. 2022.
\newblock User satisfaction estimation with sequential dialogue act modeling in
  goal-oriented conversational systems.
\newblock In \emph{Proceedings of the ACM Web Conference 2022}, pages
  2998--3008.

\bibitem[{Deriu et~al.(2021)Deriu, Rodrigo, Otegi, Echegoyen, Rosset, Agirre,
  and Cieliebak}]{deriu2021}
Jan Deriu, Alvaro Rodrigo, Arantxa Otegi, Guillermo Echegoyen, Sophie Rosset,
  Eneko Agirre, and Mark Cieliebak. 2021.
\newblock Survey on evaluation methods for dialogue systems.
\newblock \emph{Artificial Intelligence Review}, 54(1):755--810.

\bibitem[{Eric et~al.(2019)Eric, Goel, Paul, Kumar, Sethi, Ku, Goyal, Agarwal,
  Gao, and Hakkani-Tur}]{eric2019multiwoz}
Mihail Eric, Rahul Goel, Shachi Paul, Adarsh Kumar, Abhishek Sethi, Peter Ku,
  Anuj~Kumar Goyal, Sanchit Agarwal, Shuyang Gao, and Dilek Hakkani-Tur. 2019.
\newblock Multi{WOZ} 2.1: A consolidated multi-domain dialogue dataset with
  state corrections and state tracking baselines.
\newblock \emph{arXiv preprint arXiv:1907.01669}.

\bibitem[{Ghazarian et~al.(2021)Ghazarian, Hedayatnia, Papangelis, Liu, and
  Hakkani-Tur}]{ghazarian2021user}
Sarik Ghazarian, Behnam Hedayatnia, Alexandros Papangelis, Yang Liu, and Dilek
  Hakkani-Tur. 2021.
\newblock User response and sentiment prediction for automatic dialogue
  evaluation.
\newblock \emph{arXiv preprint arXiv:2111.08808}.

\bibitem[{Ghazarian et~al.(2022{\natexlab{a}})Ghazarian, Hedayatnia,
  Papangelis, Liu, and Hakkani-Tur}]{ghazarian-etal-2022-wrong}
Sarik Ghazarian, Behnam Hedayatnia, Alexandros Papangelis, Yang Liu, and Dilek
  Hakkani-Tur. 2022{\natexlab{a}}.
\newblock \href {https://doi.org/10.18653/v1/2022.findings-acl.331} {What is
  wrong with you?: Leveraging user sentiment for automatic dialog evaluation}.
\newblock In \emph{Findings of the Association for Computational Linguistics:
  ACL 2022}, pages 4194--4204, Dublin, Ireland. Association for Computational
  Linguistics.

\bibitem[{Ghazarian et~al.(2020)Ghazarian, Weischedel, Galstyan, and
  Peng}]{ghazarian2020engagement}
Sarik Ghazarian, Ralph Weischedel, Aram Galstyan, and Nanyun Peng. 2020.
\newblock Predictive engagement: An efficient metric for automatic evaluation
  of open-domain dialogue systems.
\newblock In \emph{Proceedings of the AAAI Conference on Artificial
  Intelligence}, volume 34.05, pages 7789--7796.

\bibitem[{Ghazarian et~al.(2022{\natexlab{b}})Ghazarian, Wen, Galstyan, and
  Peng}]{ghazarian-etal-2022-deam}
Sarik Ghazarian, Nuan Wen, Aram Galstyan, and Nanyun Peng. 2022{\natexlab{b}}.
\newblock \href {https://doi.org/10.18653/v1/2022.acl-long.57} {{DEAM}:
  Dialogue coherence evaluation using {AMR}-based semantic manipulations}.
\newblock In \emph{Proceedings of the 60th Annual Meeting of the Association
  for Computational Linguistics (Volume 1: Long Papers)}, pages 771--785,
  Dublin, Ireland. Association for Computational Linguistics.

\bibitem[{Gupta et~al.(2021)Gupta, Fan, Liu, Yao, Ling, Zhou, Pham, and
  Guo}]{gupta2021robertaiq}
Saurabh Gupta, Xing Fan, Derek Liu, Benjamin Yao, Yuan Ling, Kun Zhou,
  Tuan-Hung Pham, and Edward Guo. 2021.
\newblock Ro{BERT}a{IQ}: An efficient framework for automatic interaction
  quality estimation of dialogue systems.
\newblock In \emph{Proceedings of DeMaL, Second International Workshop on
  Data-Efficient Machine Learning (KDD 2021)}.

\bibitem[{Higashinaka et~al.(2014)Higashinaka, Meguro, Imamura, Sugiyama,
  Makino, and Matsuo}]{higashinaka2014evaluate}
Ryuichiro Higashinaka, Toyomi Meguro, Kenji Imamura, Hiroaki Sugiyama, Toshiro
  Makino, and Yoshihiro Matsuo. 2014.
\newblock Evaluating coherence in open domain conversational systems.
\newblock In \emph{15th Annual Conference of the International Speech
  Communication Association ({INTERSPEECH})}, pages 130--134. {ISCA}.

\bibitem[{Huang et~al.(2020)Huang, Ye, Qin, Lin, and Liang}]{huang2020grade}
Lishan Huang, Zheng Ye, Jinghui Qin, Liang Lin, and Xiaodan Liang. 2020.
\newblock {GRADE:} automatic graph-enhanced coherence metric for evaluating
  open-domain dialogue systems.
\newblock In \emph{Proceedings of the 2020 Conference on Empirical Methods in
  Natural Language Processing, {EMNLP} 2020, Online, November 16-20, 2020},
  pages 9230--9240. Association for Computational Linguistics.

\bibitem[{Ji et~al.(2022)Ji, Graham, Jones, Lyu, and
  Liu}]{ji-etal-2022-achieving}
Tianbo Ji, Yvette Graham, Gareth Jones, Chenyang Lyu, and Qun Liu. 2022.
\newblock \href {https://doi.org/10.18653/v1/2022.acl-long.445} {Achieving
  reliable human assessment of open-domain dialogue systems}.
\newblock In \emph{Proceedings of the 60th Annual Meeting of the Association
  for Computational Linguistics (Volume 1: Long Papers)}, pages 6416--6437,
  Dublin, Ireland. Association for Computational Linguistics.

\bibitem[{Lubis et~al.(2022)Lubis, Geishauser, Lin, van Niekerk, Heck, Feng,
  and Gašić}]{lubis-etal-2022-reinforcement}
Nurul Lubis, Christian Geishauser, Hsien-Chin Lin, Carel van Niekerk, Michael
  Heck, Shutong Feng, and Milica Gašić. 2022.
\newblock Dialogue evaluation with offline reinforcement learning.
\newblock \emph{arXiv preprint arXiv:2209.00876}.

\bibitem[{Mehri and Esk{\'{e}}nazi(2020)}]{mehri2020fed}
Shikib Mehri and Maxine Esk{\'{e}}nazi. 2020.
\newblock Unsupervised evaluation of interactive dialog with {DialoGPT}.
\newblock In \emph{Proceedings of the 21th Annual Meeting of the Special
  Interest Group on Discourse and Dialogue, SIGdial 2020}, pages 225--235.
  Association for Computational Linguistics.

\bibitem[{Mesgar et~al.(2020)Mesgar, B{\"{u}}cker, and
  Gurevych}]{mesgar2020dialogue}
Mohsen Mesgar, Sebastian B{\"{u}}cker, and Iryna Gurevych. 2020.
\newblock Dialogue coherence assessment without explicit dialogue act labels.
\newblock In \emph{Proceedings of the 58th Annual Meeting of the Association
  for Computational Linguistics, {ACL} 2020, Online, July 5-10, 2020}, pages
  1439--1450. Association for Computational Linguistics.

\bibitem[{Peng et~al.(2020)Peng, Li, Li, Shayandeh, Liden, and
  Gao}]{peng2020soloist}
Baolin Peng, Chunyuan Li, Jinchao Li, Shahin Shayandeh, Lars Liden, and
  Jianfeng Gao. 2020.
\newblock Soloist: Few-shot task-oriented dialog with a single pretrained
  auto-regressive model.
\newblock \emph{arXiv preprint arXiv:2005.05298}.

\bibitem[{Rastogi et~al.(2020)Rastogi, Zang, Sunkara, Gupta, and
  Khaitan}]{rastogi2020schema}
Abhinav Rastogi, Xiaoxue Zang, Srinivas Sunkara, Raghav Gupta, and Pranav
  Khaitan. 2020.
\newblock Schema-guided dialogue state tracking task at dstc8.
\newblock \emph{arXiv preprint arXiv:2002.01359}.

\bibitem[{Schmitt and Ultes(2015)}]{schmitt2015interaction}
Alexander Schmitt and Stefan Ultes. 2015.
\newblock Interaction quality: assessing the quality of ongoing spoken dialog
  interaction by experts—and how it relates to user satisfaction.
\newblock \emph{Speech Communication}, 74:12--36.

\bibitem[{Schmitt et~al.(2012)Schmitt, Ultes, and Minker}]{schmitt2012}
Alexander Schmitt, Stefan Ultes, and Wolfgang Minker. 2012.
\newblock A parameterized and annotated spoken dialog corpus of the {CMU}
  {L}et’s {G}o bus information system.
\newblock In \emph{Proceedings of the Eighth International Conference on
  Language Resources and Evaluation (LREC'12)}, pages 3369--3373.

\bibitem[{See et~al.(2019)See, Roller, Kiela, and Weston}]{see2019}
Abigail See, Stephen Roller, Douwe Kiela, and Jason Weston. 2019.
\newblock \href {https://arxiv.org/abs/1902.08654} {What makes a good
  conversation? how controllable attributes affect human judgments}.
\newblock In \emph{North American Chapter of the Association for Computational
  Linguistics (NAACL)}.

\bibitem[{Siro et~al.(2022)Siro, Aliannejadi, and de~Rijke}]{siro}
Clemencia Siro, Mohammad Aliannejadi, and Maarten de~Rijke. 2022.
\newblock \href {https://doi.org/10.1145/3477495.3531798} {Understanding user
  satisfaction with task-oriented dialogue systems}.
\newblock In \emph{Proceedings of the 45th International ACM SIGIR Conference
  on Research and Development in Information Retrieval}.

\bibitem[{Sun et~al.(2021)Sun, Zhang, Balog, Ren, Ren, Chen, and
  de~Rijke}]{sun2021}
Weiwei Sun, Shuo Zhang, Krisztian Balog, Zhaochun Ren, Pengjie Ren, Zhumin
  Chen, and Maarten de~Rijke. 2021.
\newblock Simulating user satisfaction for the evaluation of task-oriented
  dialogue systems.
\newblock In \emph{Proceedings of the 44th International ACM SIGIR Conference
  on Research and Development in Information Retrieval}, pages 2499--2506.

\bibitem[{Ultes et~al.(2014)Ultes, ElChab, and Minker}]{ultes2014application}
Stefan Ultes, Robert ElChab, and Wolfgang Minker. 2014.
\newblock Application and evaluation of a conditioned hidden markov model for
  estimating interaction quality of spoken dialogue systems.
\newblock In \emph{Natural Interaction with Robots, Knowbots and Smartphones:
  Putting Spoken Dialog Systems into Practice}, pages 303--312. Springer.

\bibitem[{Vakulenko et~al.(2018)Vakulenko, de~Rijke, Cochez, Savenkov, and
  Polleres}]{vakulenko2018measuring}
Svitlana Vakulenko, Maarten de~Rijke, Michael Cochez, Vadim Savenkov, and Axel
  Polleres. 2018.
\newblock Measuring semantic coherence of a conversation.
\newblock In \emph{International Semantic Web Conference}, pages 634--651.

\bibitem[{Walker et~al.(2000)Walker, Kamm, and Litman}]{walker2000towards}
Marilyn Walker, Candace Kamm, and Diane Litman. 2000.
\newblock Towards developing general models of usability with {PARADISE}.
\newblock \emph{Natural Language Engineering}, 6(3-4):363--377.

\bibitem[{Walker et~al.(1997)Walker, Litman, Kamm, and
  Abella}]{walker1997paradise}
Marilyn Walker, Diane Litman, Candace Kamm, and Alicia Abella. 1997.
\newblock {PARADISE}: A framework for evaluating spoken dialogue agents.
\newblock In \emph{Proceedings of the 35th Annual Meeting of the ACL and Eighth
  Conference of the European Chapter of the Association for Computational
  Linguistics}, page 271–280.

\bibitem[{Wu et~al.(2020)Wu, Hoi, Socher, and Xiong}]{wu2020tod}
Chien-Sheng Wu, Steven Hoi, Richard Socher, and Caiming Xiong. 2020.
\newblock {TOD}-{BERT}: Pre-trained natural language understanding for
  task-oriented dialogue.
\newblock \emph{arXiv preprint arXiv:2004.06871}.

\bibitem[{Yang et~al.(2021)Yang, Li, and Quan}]{yang2021ubar}
Yunyi Yang, Yunhao Li, and Xiaojun Quan. 2021.
\newblock Ubar: Towards fully end-to-end task-oriented dialog system with
  {GPT}-2.
\newblock In \emph{Proceedings of the AAAI Conference on Artificial
  Intelligence}, volume 35.16, pages 14230--14238.

\bibitem[{Yeh et~al.(2021)Yeh, Eskenazi, and Mehri}]{yeh2021comprehensive}
Yi-Ting Yeh, Maxine Eskenazi, and Shikib Mehri. 2021.
\newblock A comprehensive assessment of dialog evaluation metrics.
\newblock \emph{arXiv preprint arXiv:2106.03706}.

\bibitem[{Zhang et~al.(2021)Zhang, Chen, D'Haro, Zhang, Friedrichs, Lee, and
  Li}]{zheng2021dynaeval}
Chen Zhang, Yiming Chen, Luis~Fernando D'Haro, Yan Zhang, Thomas Friedrichs,
  Grandee Lee, and Haizhou Li. 2021.
\newblock {DynaEval}: Unifying turn and dialogue level evaluation.
\newblock In \emph{Proceedings of the 59th Annual Meeting of the Association
  for Computational Linguistics and the 11th International Joint Conference on
  Natural Language Processing}, pages 5676--5689. Association for Computational
  Linguistics.

\end{thebibliography}
\bibliographystyle{acl_natbib}

\appendix
\clearpage
\section{Dialog Quality Annotation Workflow Design}
\label{sec:appendix}

Here are some selected questions for collecting human annotations used in the DQA workflow. The design of this workflow was inspired by \citet{see2019}. In each annotation task, a multi-turn dialog is presented to the data annotator (DA) in its entirety. The dialog consists of a sequence of turns. Each turn consists of a User request and a System response.

\noindent \newline
{\bf Turn Level:}
First, the DA is asked to rate every turn in the dialog.

\noindent \newline
{\bf Provide an overall rating for the System’s response in the current turn}
\begin{tasks}[style=itemize, label-align=left, label-offset={0mm}](1)%
	\task 1-Terrible
	\task 2-Bad
	\task 3-Ok
	\task 4-Good
	\task 5-Excellent
\end{tasks}

\noindent \newline
{\bf Dialog Level:}
Next the DA is asked to answer a series of dialog-level questions to capture the overall rating along with some salient attributes of the dialog.

\noindent \newline
{\bf [User Satisfaction] Rate the overall user satisfaction based on their interaction in the dialog}
\begin{tasks}[style=itemize, label-align=left, label-offset={0mm}](1)%
	\task 1-Very Dissatisfied
	\task 2-Dissatisfied
	\task 3-Normal
	\task 4-Satisfied
	\task 5-Very Satisfied
\end{tasks}

\noindent \newline
{\bf [Goal Completion] How many goals are in the dialog?}
\begin{tasks}[style=itemize, label-align=left, label-offset={0mm}](1)%
	\task Zero
	\task One
	\task Many
\end{tasks}

\noindent \newline
{\bf [Goal Progression] Did the user make progress towards achieving their goals?}
\begin{tasks}[style=itemize, label-align=left, label-offset={0mm}](1)%
	\task No Progress
	\task Some Progress
	\task Full Progress
\end{tasks}

\noindent \newline
{\bf [Goal Completion] How many goals did the user complete in the dialog?}
\begin{tasks}[style=itemize, label-align=left, label-offset={0mm}](1)%
	\task None Completed
	\task Some Completed
	\task All Completed
\end{tasks}

\noindent \newline
{\bf [Goal Friction] Did the user encounter friction trying to achieve their goals in the dialog?}
\begin{tasks}[style=itemize, label-align=left, label-offset={0mm}](1)%
	\task Lots of Friction
	\task Some Friction
	\task No Friction
\end{tasks}

\noindent \newline
{\bf [Coherence] How often did the System say something which did NOT make sense?}
\begin{tasks}[style=itemize, label-align=left, label-offset={0mm}](1)%
	\task Never Made Sense
	\task Some Made Sense
	\task All Made Sense
\end{tasks}

\noindent \newline
{\bf [Sentiment] Describe the user’s sentiment in the conversation with the System}
\begin{tasks}[style=itemize, label-align=left, label-offset={0mm}](1)%
	\task Negative
	\task Neutral
	\task Positive
\end{tasks}
\clearpage
\section{Dialog patterns}
In this section, we compare the behavior of the baselines methods and DQM model predictions on specific customer interaction patterns found in multi-turn conversations. Note that the dialog samples in this appendix are synthetic examples fabricated to illustrate common use cases supported by our platform.

\subsection{Fatal Turns in a Dialog}
\label{sec:fatal-turns}
Sometimes a dialog can be considered defective based on the interaction in a single turn. We refer to such turns as ``fatal turns'' (See Turn 2 in Figure~\ref{fig:dialog-example-fatal-turn}). This example illustrates that baseline methods which aggregate turn-level information do not adequately capture the non-linear nature of the overall customer satisfaction in a multi-turn dialog. In contrast, we observed that DQM has a higher chance of detecting defective dialogs with fatal turns. 

\begin{figure}[ht!]
	\centering
	\scriptsize
	\resizebox{7.5cm}{!}{%
		\begin{tabular}{p{0.35cm}|p{2cm}|p{3.5cm}|p{0.5cm}}
			\hline
			\textbf{Turn} & \textbf{User utterance}  & \textbf{System response}                                                                                                                                                  & \textbf{TLD score} \\ \hline
			1             & where is my package      & 2 packages for your order should arrive today by 10:00 PM. One containing toothbrush. Another package should arrive tomorrow. Containing Smartwatch and Wireless Speaker & 0.05                \\ 
			2             & what's in my package  & I am having trouble accessing your information. Try again later & 0.75               \\ 
			3             & notify me when they come     & Your package containing toothbrush is arriving by 10:00 PM.                                                                                                               & 0.01               \\ \hline
			\multicolumn{3}{l|}{Dialog level ratings} & Rating \\ \hline
			\multicolumn{3}{l|}{Mean TLD} & 0.27 \\ 
			\multicolumn{3}{l|}{Last-turn TLD} & 0.01 \\
			\multicolumn{3}{l|}{TLD-U} & 0.27 \\ 
			\multicolumn{3}{l|}{DQM} & 0.80 \\
			\multicolumn{3}{l|}{Human Rating from DQA Workflow} & 2 \\  \hline
		\end{tabular}}
	\caption{Dialog with a fatal turn (Turn 2)}.
	\label{fig:dialog-example-fatal-turn}
\end{figure}

\subsection{User Rephrase}
\label{sec:user-rephrase}
Another common dialog pattern is a user-rephrase (See Figure~\ref{fig:dialog-example-user-rephrase}). Here the user simply repeats or slightly rephrases their initial request hoping for a better system response. User rephrases in most cases indicate customer friction and often get poor dialog-level ratings. Since the TLD model we use for encoding turns (see \S~\ref{sec:DQM-model}) considers context around the current turn, it learns a strong association between user rephrases and defective turns.  

\begin{figure}[ht!]
	\centering
	\scriptsize
	\resizebox{7.5cm}{!}{%
		\begin{tabular}{p{0.35cm}|p{2cm}|p{3.5cm}|p{0.5cm}}
			\hline
			\textbf{Turn} & \textbf{User utterance}               & \textbf{System response}                                            & \textbf{TLD score} \\ \hline
			1             & when is the delivery gonna be here    & Sorry, I don’t have an answer for that                              & 0.99               \\ 
			2             & when is my delivery going to get here & Sorry, I don’t have an answer for that                              & 0.99               \\ 
			3             & when will my package arrive           & Your toothbrush, shampoo and 3 more items should arrive by Friday & 0.02               \\ \hline
			\multicolumn{3}{l|}{Dialog level ratings} & Rating \\ \hline
			\multicolumn{3}{l|}{Mean TLD} & 0.66 \\ 
			\multicolumn{3}{l|}{Last-turn TLD} & 0.02 \\
			\multicolumn{3}{l|}{TLD-U} & 0.66 \\ 
			\multicolumn{3}{l|}{DQM} & 0.74 \\
			\multicolumn{3}{l|}{Human Rating from DQA Workflow} & 2 \\  \hline
		\end{tabular}}
	\caption{Dialog with a rephrased user request}
	\label{fig:dialog-example-user-rephrase}
\end{figure}

\subsection{User Refinement}
\label{sec:user-refinements}
In use cases involving topic exploration or navigation of recommendations, a user progressively adds more constraints to refine their earlier requests (see Figure~\ref{fig:dialog-example-user-refinement}). Unlike the user rephrase pattern described in Section~\ref{sec:user-rephrase}, the previous turns leading to more refinement do not necessarily indicate a unsatisfactory customer experience. A well performing dialog-quality model should learn to distinguish between frictional rephrases and non-frictional refinements.  We note that TLD-based baselines have a bias towards incorrectly predicting refinements as defects. As DQM uses TLD as an input signal, DQM too struggles to effectively distinguish between frictional rephrase and non-frictional refinement.  
\begin{figure}[ht!]
	\centering
	\scriptsize
	\resizebox{7.5cm}{!}{%
		\begin{tabular}{p{0.35cm}|p{2cm}|p{3.5cm}|p{0.5cm}}
			\hline
			\textbf{Turn} & \textbf{User utterance}         & \textbf{System response}           & \textbf{TLD score} \\ \hline
			1             & how much does milk cost         & a gallon of milk costs \$6         & 0.90               \\     
			2             & how much does organic milk cost & a gallon of organic milk costs \$7 & 0.20               \\  \hline   
			\multicolumn{3}{l|}{Dialog level ratings} & Rating \\ \hline
			\multicolumn{3}{l|}{Mean TLD} & 0.55 \\ 
			\multicolumn{3}{l|}{Last-turn TLD} & 0.20 \\
			\multicolumn{3}{l|}{TLD-U} & 0.55 \\ 
			\multicolumn{3}{l|}{DQM} & 0.63 \\
			\multicolumn{3}{l|}{Human Rating from DQA Workflow} & 5 \\  \hline
			        
		\end{tabular}}
	\caption{Dialog with a user query refinement}
	\label{fig:dialog-example-user-refinement}
\end{figure}

\end{document}